\documentclass[letterpaper]{article} 
\usepackage{aaai2027}  
\usepackage[hyphens]{url}  
\usepackage{graphicx} 
\usepackage{natbib}  
\usepackage{caption} 
\usepackage{algorithm}
\usepackage{algorithmic}
\usepackage{amsmath}
\usepackage{amssymb}
\usepackage{graphicx}
\usepackage{caption}
\usepackage{newfloat}
\usepackage{listings}
\DeclareCaptionStyle{ruled}{labelfont=normalfont,labelsep=colon,strut=off} 
\floatstyle{ruled}
\newfloat{listing}{tb}{lst}{}
\floatname{listing}{Listing}

\usepackage{booktabs}

\nocopyright

\title{TransAnyText: Translating Arbitrary Text in E-commerce Images\\ via Structured Visual Generation}
\author{%
  \textbf{Xiaoan Liu}$^{1,2,\ast}$,
  \textbf{Lichen Ma}$^{2,3,\ast}$,
  \textbf{Zipeng Guo}$^{2}$,
  \textbf{Yu He}$^{2}$,
  \textbf{Xiaoyan Su}$^{4}$,
  \textbf{Shaojie Guo}$^{2}$,
  \textbf{Hao Yang}$^{3}$,\\
  \textbf{Jingling Fu}$^{2}$,
  \textbf{Xiaolong Fu}$^{2}$,
  \textbf{Zhen Chen}$^{2}$,
  \textbf{Yu Guo}$^{3}$,
  \textbf{Fei Wang}$^{3}$,
  \textbf{Xinyi Liu}$^{1}$,
  \textbf{Yongjun Zhang}$^{1,\ddagger}$,\\
  \textbf{Ke Zhang}$^{2}$,
  \textbf{Junshi Huang}$^{2,\dagger}$
}
\affiliations{%
  $^1$Wuhan University,   
  $^2$JD.com,
  $^3$State Key Laboratory of Human-Machine Hybrid Augmented Intelligence, Institute of Artificial Intelligence and Robotics, Xi'an Jiaotong University,
  $^4$The Hong Kong University of Science and Technology (Guangzhou)
}
\makeatletter
\g@addto@macro\@thanks{%
  \footnotetext[1]{Equal contribution.}%
  \footnotetext[2]{Project Lead.}%
  \footnotetext[3]{Corresponding Author.}%
}
\makeatother

\begin{document}

\maketitle

\begin{abstract}
Cross-border e-commerce image translation is essential for global retail, where product images, banners, and detail pages need to be produced in different languages. Existing methods struggle to achieve accurate translation, faithful visual identity preservation, and easy-to-edit outputs, simultaneously. To address these challenges, we introduce TransAnyText, a structured visual code framework that reformulates image text translation as generating renderable HTML patches from source images and target languages. Our framework decouples semantic generation from pixel rendering: a vision-language model~(VLM) handles visual understanding, cross-lingual translation, and structured visual generation, while a diffusion model performs background inpainting and pixel-level refinement, followed by deterministic rendering to synthesize the final image. Based on this formulation, we develop a three-stage post-training framework, where supervised fine-tuning (SFT) establishes the image-to-code mapping, privilege-gap weighted self-distillation (PWSD) improves the learning of style and layout tokens, and reinforcement learning with verifiable rewards (RLVR) further optimizes task-level performance. We further introduce TransAnyDataset and TransAnyBench, a multilingual dataset and benchmark for e-commerce image translation. Extensive experiments demonstrate competitive performance against cascaded pipelines, open-source end-to-end models, and closed-source image editing systems, providing an effective, controllable, and editable solution for cross-border e-commerce image translation.
\end{abstract}

\section{Introduction}

\begin{figure}[!t]
    \centering
    \includegraphics[width=0.94\linewidth]{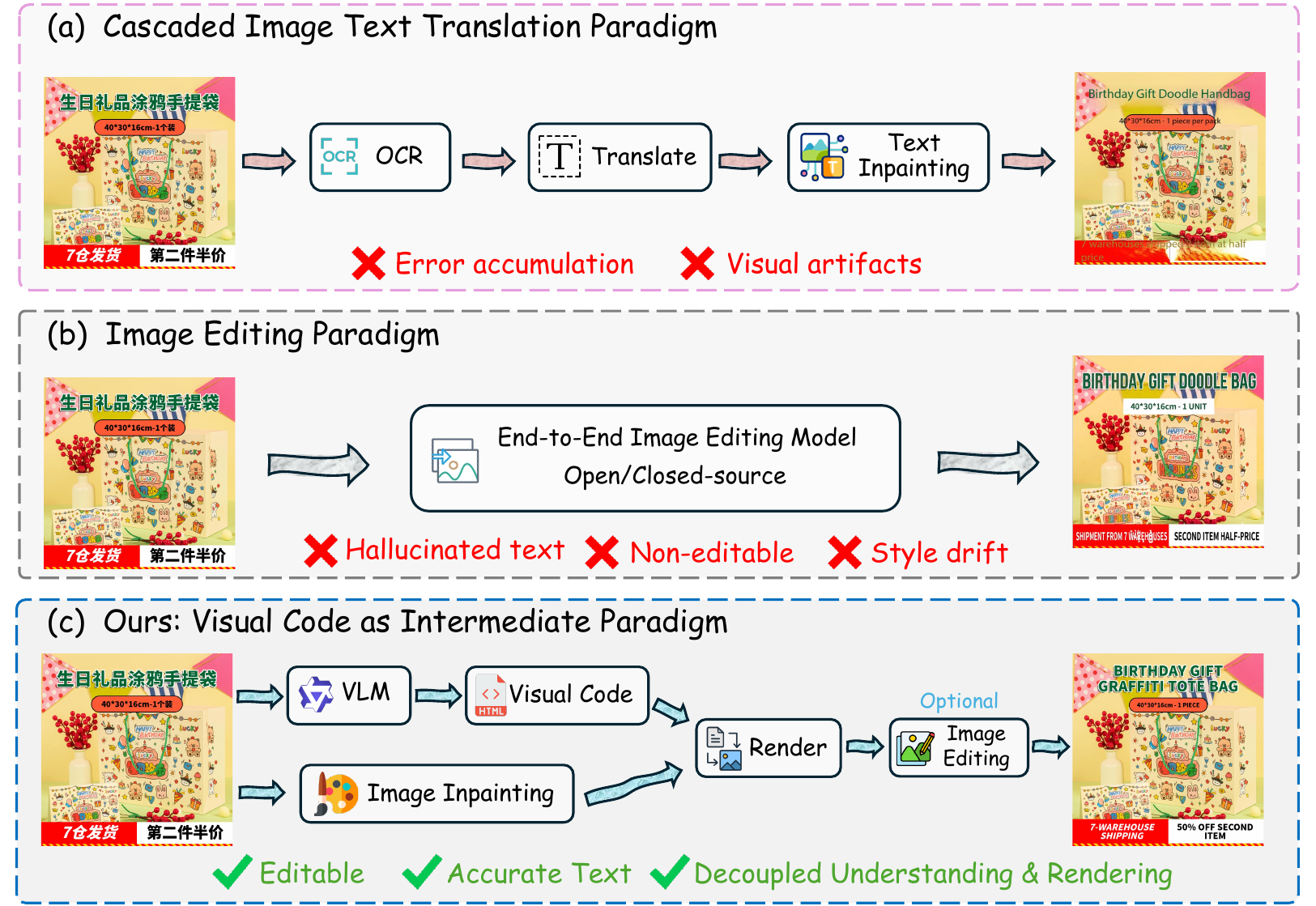}
    \caption{Comparison of paradigms: (a) the cascaded pipeline of detection, translation, erasing, and re-rendering, which suffers from error accumulation and visual artifacts; (b) the end-to-end pixel edit paradigm, which jointly performs translation and rendering but often hallucinates or distorts text; and (c) our structured visual code paradigm, which decouples semantic generation from pixel rendering to yield accurate, editable, and visually faithful results.}
    \label{fig:paradigm}
    \vspace{-2mm}
\end{figure}

\begin{figure*}[!t]
    \centering
    \includegraphics[width=0.95\textwidth]{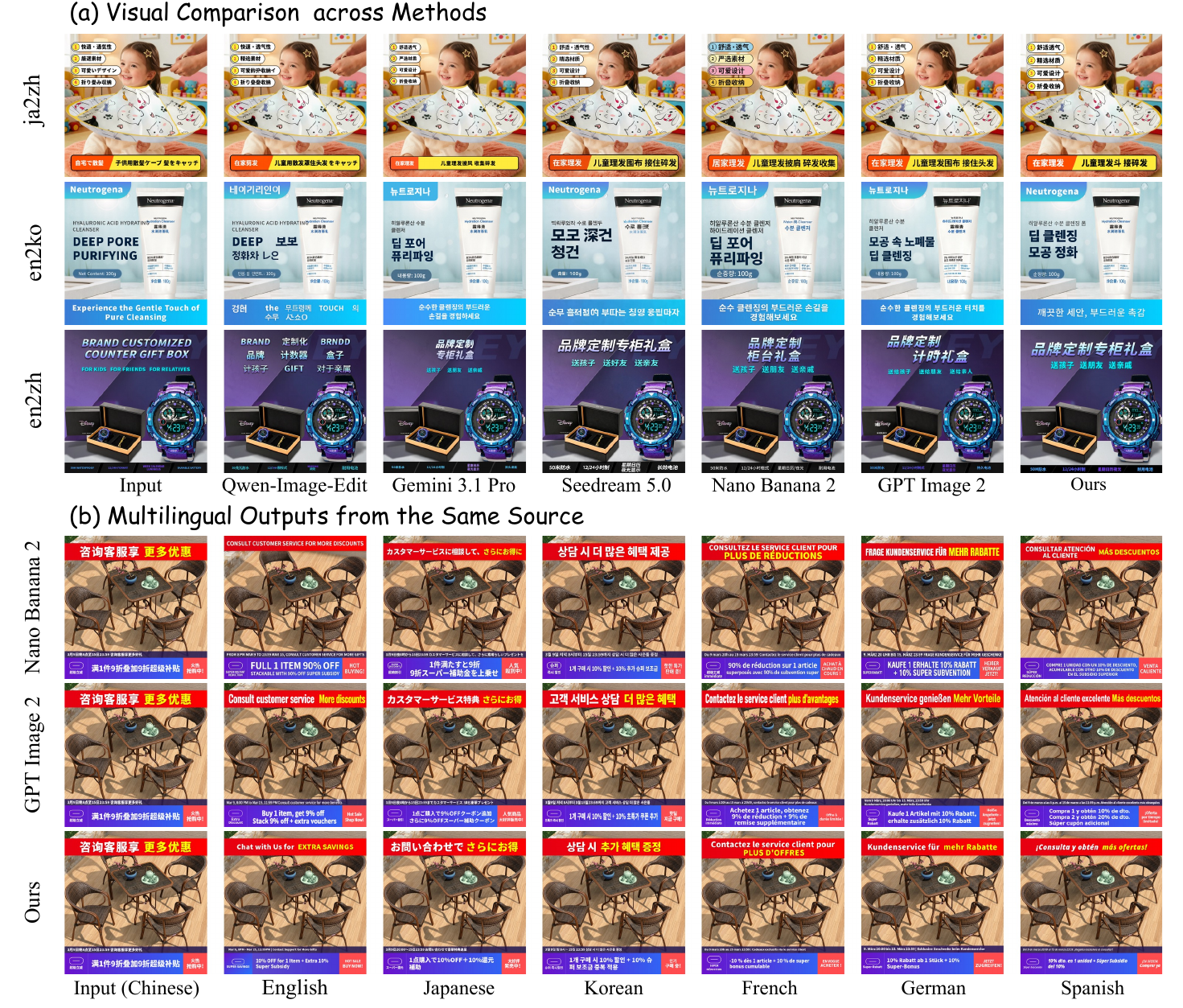}
    \vspace{-1mm}
    \caption{Visual comparison on e-commerce image text translation. (a) Qualitative comparison with existing methods; (b) qualitative comparison with closed-source models on multilingual outputs from a single Chinese source image.}
    \vspace{-2mm}
    \label{fig:teaser}
\end{figure*}

Cross-border e-commerce has become a key growth driver for global retail platforms, where product images, banners, and detail pages are essential for product presentation and customer conversion~\cite{gao2025postermaker,fan2026autopp,qin2026innoads,guo2026gmo}. These visual assets require frequent localization across markets and languages. However, large-scale image localization still relies on substantial human intervention, leading to high costs and low scalability. More importantly, this task goes beyond simply translating text into different languages. A successful solution must satisfy three key requirements: 1) adapting layouts to accommodate language-dependent text variations; 2) preserving visual identity elements such as brand fonts, badges, and decorative effects; and 3) maintaining editability for downstream operations. Developing an open, controllable, and editable solution for multilingual e-commerce image translation remains a critical challenge.

Existing in-image translation methods mainly follow two paradigms. The cascaded route in Figure~\ref{fig:paradigm}~(a) decomposes the task into text extraction, translation, and rendering, enabling each stage to leverage specialized models~\cite{li2024first, tuo2024anytext, zeng2024textctrl}. However, errors accumulate along the pipeline, erasing and re-rendering text often introduces artifacts on complex backgrounds~\cite{shu2025visual}. The end-to-end pixel route in Figure~\ref{fig:paradigm}~(b)  directly generates target-language images~\cite{lan2024translatotron, wu2025qwenimage, lyu2026unitranslator}, but requires the model to jointly handle visual understanding, cross-lingual translation, and text rendering. This leads to hallucinated, missing, or incorrect text, while scaling to multilingual scenarios requires substantial data and training resources.  Although closed-source image editing models achieve strong visual quality, their pixel-based outputs remain difficult to control and edit, due to high API costs, data compliance concerns, and limited fine-tuning flexibility. These limitations highlight a fundamental gap: pixel-based representations struggle to explicitly encode the structural and visual constraints required for e-commerce image translation, such as adaptive layout, text fidelity, style consistency, and easy editability.

To tackle these limitations, we introduce structured visual code to decouple semantic generation from pixel rendering, as illustrated in Figure~\ref{fig:paradigm}~(c). In this framework, a vision-language model~(VLM) handles visual understanding, cross-lingual translation, and structured visual generation, while a diffusion model specializes in background inpainting and pixel-level refinement. The final image is deterministically synthesized by a renderer from the structured representation, enabling faithful and controllable translation. Specifically, we reformulate e-commerce image translation as generating a renderable HTML $H$ from a source image $I$ and a target language $L_t$. This formulation ensures deterministic text rendering while providing explicit control over visual attributes and layout structure. Based on this formulation, we develop a three-stage post-training framework: SFT learns the image-to-code mapping and trains the diffusion model for background inpainting; PWSD provides dense supervision for under-optimized style and layout tokens; RLVR further aligns task-level performance through verifiable rewards covering spatial accuracy, translation quality, and style fidelity.

Our contributions are as follows. \textbf{(1) Task reformulation:} We reformulate e-commerce image translation using visual code as a structured, renderable intermediate representation, achieving competitive results compared with existing open-source and closed-source approaches. \textbf{(2) Post-training framework:} We propose TransAnyText, an open-source three-stage training framework that improves visual-token learning and aligns task-level performance with verifiable rewards. \textbf{(3) Multilingual benchmark:} We introduce TransAnyDataset and TransAnyBench, a multilingual dataset and benchmark with comprehensive evaluation of translation quality, visual fidelity, and image realism.

\section{Related Work}

\subsection{Image Text Translation}

In-image machine translation (IMT) translates text within images and renders it back while preserving original layouts and visual styles. Existing approaches are divided into cascaded systems and end-to-end models. Cascaded methods~\cite{qian2024anytrans,lu2026global,lyu2026imtbench} split the task into sequential stages: detection, recognition, translation, and inpainting. They extract text via vision models, translate it, and paste it back using text-editing generative models~\cite{ma2024chargen,ma2026text,tuo2024anytext,lan2025flux,guo2025repainter}. However, these systems suffer from a fundamental bottleneck: isolated modules introduce error accumulation. Since the "translation + paste-back" paradigm does not naturally integrate visual understanding and language generation, its inherent limitations cannot be addressed by improving individual components, limiting their scalability.

End-to-End Image-Informed Machine Translation (IIMT)
To mitigate the error propagation and pipeline complexity of cascaded systems, recent research has shifted toward end-to-end Image-Informed Machine Translation (IIMT). Early works like Translatotron-V~\cite{lan2024translatotron} decoupled translation from visual rendering using an intermediate text decoder for discrete token prediction, reducing pixel-level optimization complexity. To bridge the gap toward real-world applications, PRIM~\cite{tian2025prim} explored practical multilingual scenarios via VisTrans, which separately processes textual and background features to enhance stability and image integrity. 
Although theoretically bypassing cascading errors through unified generation, it suffers from fatal real-world flaws, including frequent text hallucinations, poor generalization, prohibitive computational costs, and uneditable pixel outputs.

\subsection{Structured Visual Generation}

With the rapid progress of multimodal large language models (MLLMs), visual code generation has emerged as an effective paradigm that bridges visual perception and executable programs~\cite{zhao2026beyond,ye2026genclaw,rodriguez2025starvector,liu2026designascode}. By representing visual content as editable, resolution-independent, and structurally explicit vector programs, these approaches provide superior scalability and editability compared with raster-based representations. Chat2SVG~\cite{wu2025chat2svg} leveraged LLMs to generate hierarchical SVG structures and refined details with a diffusion-based prior. OmniSVG~\cite{yang2026omnisvg} tokenized SVG commands and coordinates for autoregressive SVG generation with pretrained vision-language models. SVGBuilder~\cite{chen2025svgbuilder} built a reusable path-component library and employed CLIP with an autoregressive decoder for efficient SVG synthesis. PosterVerse~\cite{liu2026posterverse} combined LLM-based design parsing, diffusion-based background generation, and VLM-driven HTML generation for visual layout synthesis. Despite using visual code as an intermediate representation, existing methods mainly address unconstrained generation from text descriptions or design specifications. In contrast, image text translation is a constrained visual code generation problem, where the model must preserve the input image's layout, typography, and visual identity while performing cross-lingual translation and adaptive text reflow. We formulate this task as image-to-HTML visual text translation and develop a systematic framework based on open-source VLMs.

\begin{figure*}[t]
    \centering
    \includegraphics[width=\textwidth]{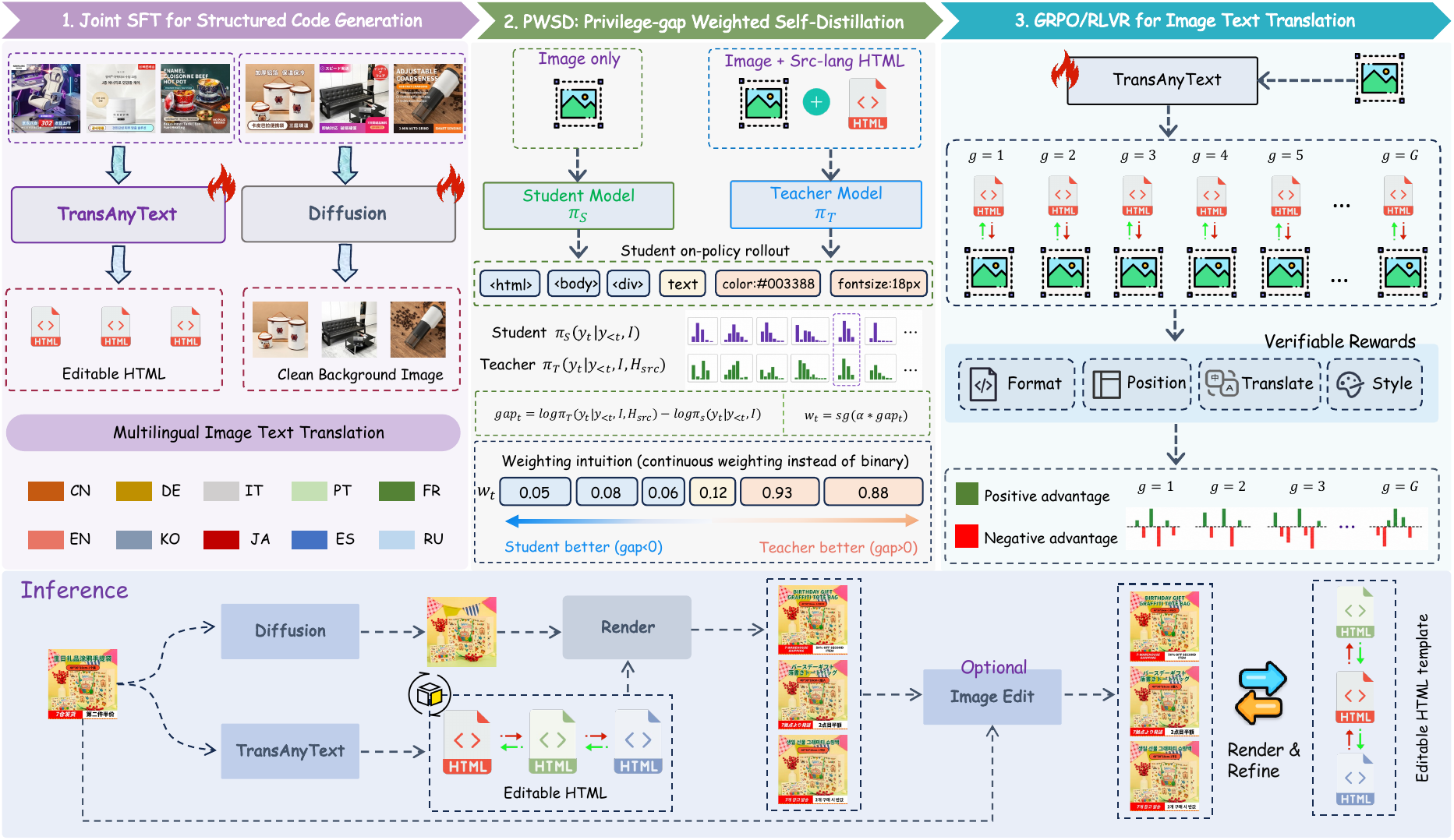}
    \vspace{-2mm}
    \caption{Overview of the TransAnyText training framework and inference pipeline. The training consists of three stages: (1) joint SFT for structured HTML code generation and background inpainting, (2) PWSD for token-level self-distillation on style and layout tokens, and (3) RLVR for task-level alignment with verifiable rewards. At inference, the VLM and diffusion model jointly produce an editable HTML and a clean background, which are deterministically rendered into the translated image.}
    \label{fig:framework}
    \vspace{-2mm}
\end{figure*}

\section{Method}

\noindent
\textbf{Overview.}
We formulate e-commerce image text translation as structured visual code generation: given a source image $I$ and a target language $L_t$, the system generates a renderable HTML patch $H$ that preserves both translation accuracy and visual identity. As shown in Figure~\ref{fig:framework}, the pipeline has four steps:
\textbf{(1) Background Inpainting:} text regions in $I$ are erased to obtain $I_{\text{bg}}$.
\textbf{(2) Structured Generation:} a VLM takes $(I, L_t)$ as input and generates $H$, encoding translated text with position, font, color, and size.
\textbf{(3) Deterministic Rendering:} $H$ is rendered onto $I_{\text{bg}}$ using Playwright to produce the translated image.
\textbf{(4) Optional Diffusion Refinement:} a diffusion model optionally refines the result to enhance visual quality. The VLM and diffusion model are initialized with SFT for image-to-HTML mapping and background inpainting, while PWSD and GRPO further optimize the VLM with token-level supervision and task-level rewards.

\subsection{Stage 1: Supervised Fine-Tuning}

SFT jointly optimizes the two modules, including the VLM-based structured generator and the background inpainting diffusion model, to establish the fundamental image-to-HTML generation capability.

\noindent\textbf{VLM.} Given paired samples $(I, H^*)$, where $H^* = [y_1, y_2, ..., y_n]$ denotes the ground-truth HTML patch represented as a sequence of tokens, we fine-tune the VLM using the standard autoregressive next-token prediction objective:
\begin{equation}
\mathcal{L}_{\text{SFT}} = -\sum_{t} \log \pi_\theta(y_t \mid y_{<t}, I, L_t)
\end{equation}

\noindent\textbf{Inpainting Diffusion Model.} We train a conditional flow matching model to learn the background inpainting process from the original image $I$ to the clean background $I_{\text{bg}}$. Specifically, a linear probability path $x_t = (1-t)\,I_{\text{bg}} + t\,I$ is constructed, and the model is optimized to regress the target velocity field $u_t = I - I_{\text{bg}}$:
\begin{equation}
\mathcal{L}_{\text{FM}} = \mathbb{E}_{t,\,(I,\,I_{\text{bg}})}\bigl\|\,v_\phi(x_t, t \mid I) - u_t\,\bigr\|_2^2
\end{equation}
The learned inpainting model provides a clean background canvas for subsequent rendering.

\subsection{Stage 2: Privilege-Gap Weighted Self Distillation}

\textbf{Motivation.} After SFT, the VLM can generate valid HTML but struggles with accurate style and layout reproduction. Style-related tokens suffer from weak structural dependency and teacher-forcing bias, leading to fragile optimization. PWSD provides dense, adaptively weighted token-level supervision before RLVR for robust initialization.

\noindent\textbf{Mechanism.} We employ the same SFT model under two different conditioning settings: a privileged teacher $\pi_T$ receives both the image $I$ and extracted source-language HTML $H_{\text{src}}$, while the student $\pi_S$ only accesses the image $I$. Given an on-policy rollout $y \sim \pi_S(\cdot \mid I)$, we measure the token-level privilege gap:
\begin{equation}
\text{gap}_t = \log \pi_T(y_t \mid y_{<t}, I, H_{\text{src}}) - \log \pi_S(y_t \mid y_{<t}, I)
\end{equation}
A stop-gradient sigmoid transformation converts the gap into an adaptive supervision weight:
\begin{equation}
w_t = \operatorname{sg}\bigl[\sigma(\alpha \cdot \text{gap}_t)\bigr]
\end{equation}
The weighted reverse KL distillation objective is then applied at the token level:
\begin{equation}
\mathcal{L}_{\text{PWSD}} = \mathbb{E}_{y \sim \pi_S}\!\Bigl[\textstyle\sum\nolimits_t w_t \operatorname{KL}(\pi_S^t \| \pi_T^t)\Bigr]
\end{equation}
This adaptive weighting mechanism emphasizes tokens with larger privilege gaps, typically corresponding to style and layout attributes, while suppressing redundant supervision on tokens where the student already agrees with the teacher.

\begin{figure*}[t]
    \centering
    \includegraphics[width=\textwidth]{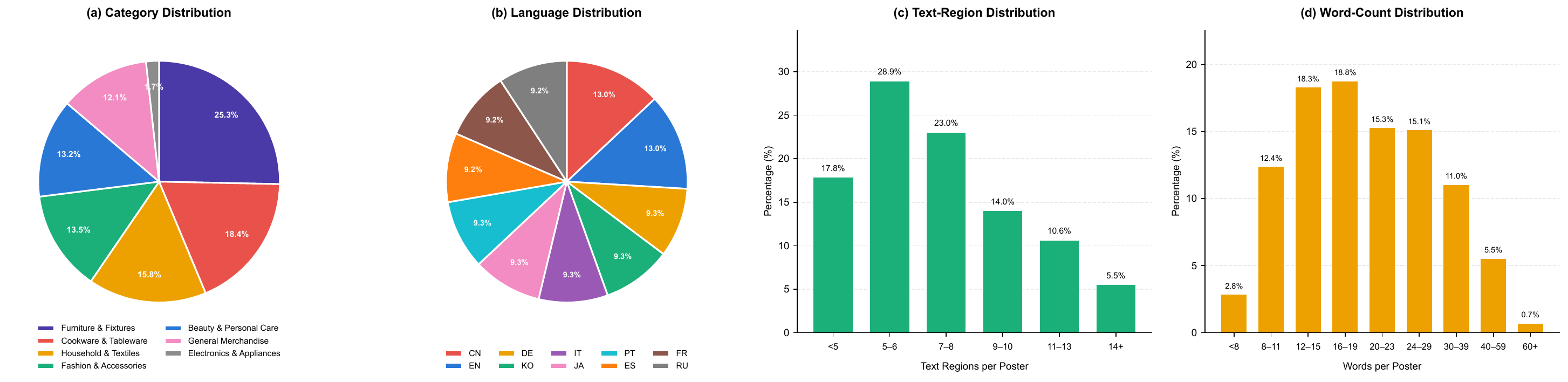}
    \vspace{-2mm}
    \caption{Dataset statistics of TransAnyDataset, including distributions over categories, languages, text regions, and word counts.}
    \label{fig:dataset}
    \vspace{-1mm}
\end{figure*}

\subsection{Stage 3: Group Relative Policy Optimization}

While PWSD provides effective token-level style guidance, further task-level optimization is required to improve overall generation performance. Therefore, we further introduce GRPO~\cite{shao2024deepseekmath} with patch-level verifiable rewards to directly optimize task-level objectives. For each input, we sample $G$ rollouts $\{y^{(g)}\}_{g=1}^{G}$ from the current policy, assign each rollout a composite reward $r^{(g)} = \sum_k \lambda_k r_k^{(g)}$, and compute group-relative advantages:
\begin{equation}
A^{(g)} = \frac{r^{(g)} - \bar{r}}{\text{std}(r)}
\end{equation}

\noindent
The GRPO objective maximizes advantage-weighted log-probability while constraining the policy deviation from the PWSD checkpoint $\pi_{\text{ref}}$:
\begin{equation}
\mathcal{J}_{\text{GRPO}} = \mathbb{E}\!\left[\sum_g A^{(g)} \log \pi_\theta(y^{(g)}) - \beta \, \text{KL}(\pi_\theta \| \pi_{\text{ref}})\right]
\end{equation}

\noindent
The overall reward is composed of four automatically verifiable components:
\begin{itemize}
    \item $\boldsymbol{r_{\text{format}}}$: verifies whether the generated HTML is syntactically valid and renderable.
    \item $\boldsymbol{r_{\text{position}}}$: measures patch-level spatial alignment between the generated layout and the source image.
    \item $\boldsymbol{r_{\text{translation}}}$: evaluates semantic correctness, linguistic fluency, and domain-specific terminology consistency.
    \item $\boldsymbol{r_{\text{style}}}$: measures the consistency of visual attributes, including color, font size, and font weight.
\end{itemize}

\subsection{Optional Diffusion Refinement}

Code-driven rendering is inherently limited in capturing pixel-level visual details. To address this issue, we introduce an open-source diffusion model as an inference-time refinement module. Conditioned on $R(H, I_{\text{bg}})$, it performs low-strength editing to reduce artifacts and enhance visual fidelity while preserving text content and layout. This design enables each module to focus on its strengths: the VLM handles semantic and structural generation, while the diffusion model specializes in pixel-level visual refinement.

\section{Dataset and Benchmark}

We introduce \textbf{TransAnyDataset} and \textbf{TransAnyBench}, the first multilingual dataset and benchmark for e-commerce image text translation. It supports \textbf{10 languages} across five writing systems and establishes a unified evaluation protocol for translation accuracy, visual fidelity, and image realism.

\noindent\textbf{Data Source and Curation.}
Source images are collected from e-commerce platforms, including product photos, banners, and promotional posters across categories such as furniture, household goods, kitchenware, fashion, beauty, and electronics. We construct multilingual samples through a four-stage pipeline: \textbf{(1) Extraction}: Gemini-3.1 Pro~\cite{gemini2026pro3} extracts structured HTML patches containing text, bounding boxes, and visual attributes.\textbf{(2) Filtering} : Qwen3.5-27B~\cite{team2026qwen35} filters out samples with incorrect text recognition or inaccurate layout alignment.\textbf{(3) Translation} : Gemini-3.1 Pro translates the samples into 8 additional languages using Chinese and English as hub languages, with e-commerce-aware constraints to preserve terminology and layout compatibility.\textbf{(4) Quality Assurance} : Qwen3-VL-32B~\cite{bai2025qwen3} evaluates each sample, and low-quality instances are discarded after human review.

\noindent\textbf{Language Coverage and Density.}
TransAnyDataset covers 10 languages, including English, Chinese, Japanese, Spanish, French, German, Portuguese, Korean, Russian, and Italian, spanning 5 writing systems. As shown in Figure~\ref{fig:dataset}, this diversity introduces challenges in cross-lingual image text translation, including script variation, text length changes, and typography adaptation. The dataset focuses on short-text, multi-region layouts commonly found in e-commerce images, while also incorporating high-density promotional banners to capture diverse layout complexities.

\noindent\textbf{Evaluation Protocol.}
Image text translation requires joint evaluation of linguistic quality and visual quality. We therefore design four complementary metrics covering translation accuracy, visual preservation, and image realism:

\begin{itemize}
\item \textbf{COMET}~\cite{rei2020comet}: A reference-based translation metric for evaluating the quality of extracted text.
\item \textbf{Translation Quality Evaluator}: A VLM-based evaluator that assesses translation accuracy, naturalness, terminology consistency, and contextual appropriateness.
\item \textbf{Visual Fidelity Evaluator}: A VLM-based evaluator that compares source and generated images to measure the preservation of visual identity, including fonts, colors, decorative elements, and product appearance.
\item \textbf{Image Realism Evaluator}: A VLM-based evaluator that assesses standalone image quality, including readability, visual coherence, and rendering artifacts.
\end{itemize}

\section{Experiments}

\begin{figure*}[!t]
    \centering
    \includegraphics[width=0.945\textwidth]{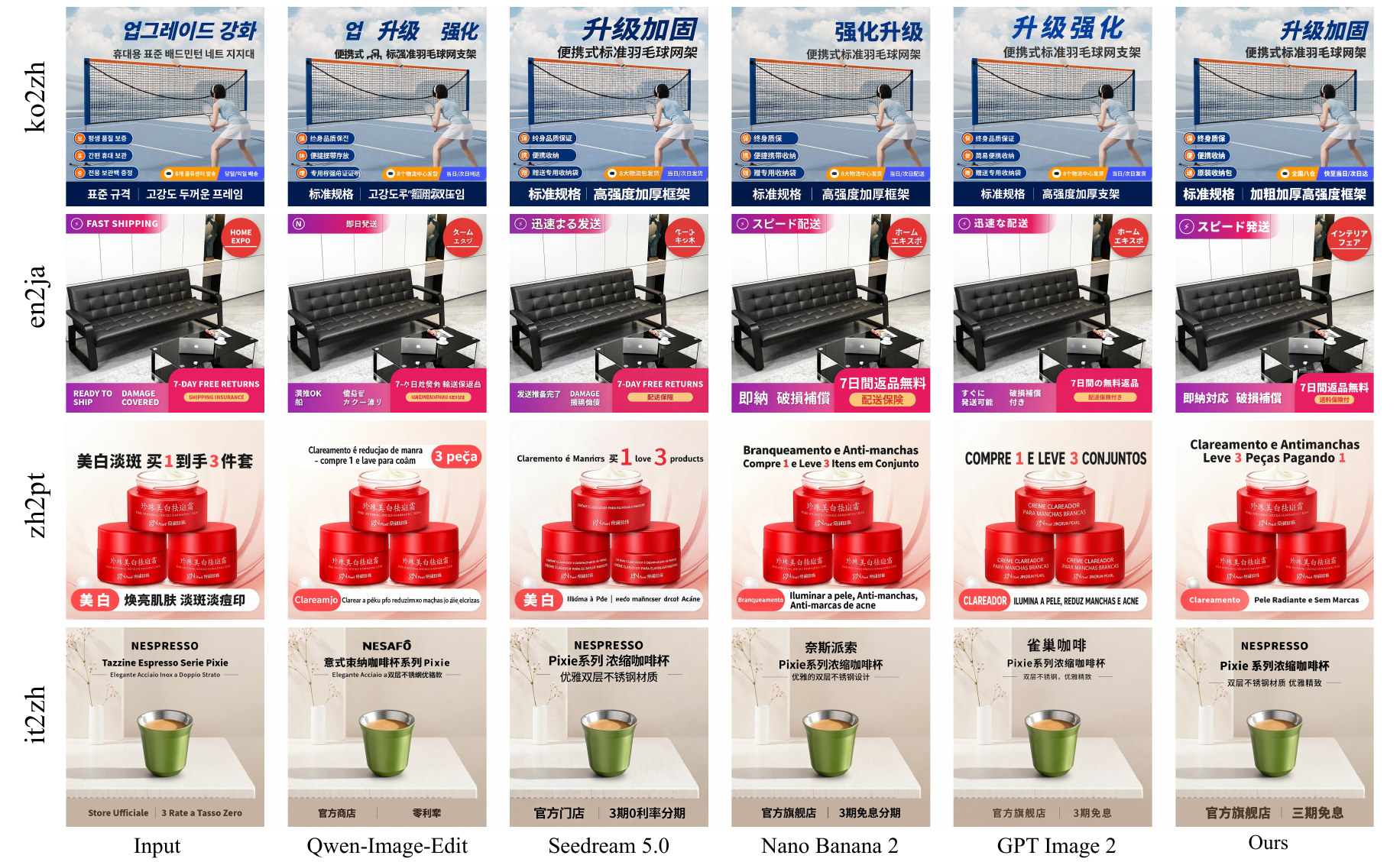}
    \caption{Visual comparison of TransAnyText with existing methods.}
    \vspace{-1mm}
    \label{fig:visual_compare}
\end{figure*}

\begin{table*}[h]
\centering
\resizebox{\textwidth}{!}{
\begin{tabular}{l cccc cccc cccc cccc}
\toprule
& \multicolumn{4}{c}{\textbf{Any$\to$Zh}} & \multicolumn{4}{c}{\textbf{Zh$\to$Any}} & \multicolumn{4}{c}{\textbf{Any$\to$En}} & \multicolumn{4}{c}{\textbf{En$\to$Any}} \\
\cmidrule(lr){2-5} \cmidrule(lr){6-9} \cmidrule(lr){10-13} \cmidrule(lr){14-17}
\textbf{Method} & COMET & T.Q. & V.F. & I.R. & COMET & T.Q. & V.F. & I.R. & COMET & T.Q. & V.F. & I.R. & COMET & T.Q. & V.F. & I.R. \\
\midrule
\multicolumn{17}{l}{\textit{Closed-source Image Editing Methods}} \\
\midrule
Seedream 5.0 & 0.794 & 9.60 &  \underline{9.52} & 9.03 & 0.440 & 3.73 & 8.11 & 5.57 & 0.528 & 5.41 & 8.41 & 6.02 & 0.549 & 5.96 & 8.59 & 6.87 \\
Nano Banana 2 & 0.782 & \underline{9.69} & \underline{9.52} & 9.22 & 0.584 & 8.73 & \underline{9.31} & 8.98 & 0.627 & 8.91 & \underline{9.38} & 8.69 & 0.632 & 9.41 & \underline{9.41} & \underline{9.22} \\
GPT Image 2 & 0.793 & 9.54 & 9.24 & \textbf{9.51} & 0.596 & 8.46 & 9.20 & \textbf{9.22} & 0.632 & 9.05 & 9.26 & \underline{9.10} & \underline{0.642} & \underline{9.45} & 9.33 & \textbf{9.33} \\
\midrule
\multicolumn{17}{l}{\textit{Cascaded Methods}} \\
\midrule
OCR+MT+T2I & 0.613 & 6.72 & 4.58 & 4.05 & 0.457 & 4.05 & 3.74 & 4.13 & 0.482 & 6.71 & 6.28 & 4.10 & 0.441 & 3.87 & 5.25 & 4.17 \\
Qwen3-VL-8B+T2I & 0.527 & 5.82 & 5.23 & 4.91 & 0.415 & 3.93 & 4.01 & 4.57 & 0.441 & 6.27 & 6.83 & 4.82 & 0.415 & 3.52 & 5.82 & 4.92 \\
\midrule
\multicolumn{17}{l}{\textit{Open-source Image Editing Methods}} \\
\midrule
FireRed-Image-Edit & 0.265 & 1.35 & 4.18 & 2.26 & 0.242 & 1.31 & 3.18 & 2.29 & 0.265 & 1.21 & 3.39 & 2.38 & 0.243 & 1.26 & 3.37 & 2.40 \\
LongCat-Image-Edit & 0.427 & 2.16 & 7.63 & 7.38 & 0.501 & 2.84 & 6.64 & 7.59 & 0.450 & 2.39 & 6.79 & 7.07 & 0.382 & 1.76 & 7.35 & 6.66 \\
Qwen-Image-Edit & 0.355 & 1.78 & 6.80 & 4.78 & 0.391 & 1.89 & 6.35 & 4.66 & 0.371 & 1.60 & 6.29 & 4.47 & 0.314 & 1.59 & 6.67 & 4.38 \\
\midrule
\multicolumn{17}{l}{\textit{Code-Driven Methods}} \\
\midrule
Qwen3.5-27B & 0.606 & 7.02 & 4.47 & 7.41 & 0.484 & 6.67 & 4.17 & 7.15 & 0.514 & 7.14 & 4.36 & 7.59 & 0.496 & 6.42 & 4.47 & 7.47 \\
GPT5.5 & 0.796 & 9.61 & 9.03 & 8.90 & 0.586 & 8.63 & 8.25 & 8.10 & 0.633 & \underline{9.19} & 8.48 & 8.33 & 0.632 & 9.39 & 8.81 & 8.75 \\
Gemini-3.1 Pro & \underline{0.807} & 9.44 & 9.43 & 9.39 & \underline{0.603} & \underline{8.87} & 9.23 & 9.04 & \underline{0.638} & 9.06 & 9.17 & 9.08 & 0.635 & 9.04 & \underline{9.41} & 9.16 \\
\midrule
Ours & \textbf{0.848} & \textbf{9.70} & \textbf{9.63} & \underline{9.48} & \textbf{0.635} & \textbf{8.94} & \textbf{9.41} & \underline{9.18} & \textbf{0.672} & \textbf{9.25} & \textbf{9.46} & \textbf{9.14} & \textbf{0.700} & \textbf{9.47} & \textbf{9.65} & \textbf{9.33} \\
\bottomrule
\end{tabular}
}
\vspace{-1mm}
\caption{Quantitative results on TransAnyBench. T.Q. represents Translation Quality, V.F. represents Visual Fidelity, and I.R. represents Image Realism. "Any" represents the aggregate over all languages except the target language. Best results are in \textbf{bold}, and second-best are \underline{underlined}.}
\vspace{-2.5mm}
\label{tab:multilingual}
\end{table*}

\subsection{Experimental Setup}

\noindent\textbf{Implementation Details.}
We adopt Qwen3.5-9B~\cite{team2026qwen35} as the backbone for structured HTML generation and perform LoRA-based fine-tuning using a learning rate of $1\times10^{-4}$ and a LoRA rank of 64. The diffusion-based inpainting model is built on FLUX.2-klein-9B~\cite{blackforestlabs2025flux2klein}, with a learning rate of $1\times10^{-5}$ and a LoRA rank of 32. The refinement module is disabled during evaluation, and can be optionally enabled during deployment with editing models (e.g., SD~\cite{esser2024scaling} or FLUX~\cite{blackforestlabs2025flux,blackforestlabs2025flux2} series). More details are provided in the supplementary materials.

\noindent\textbf{Compared Methods.}
We compare against four categories of methods: 
(1) \textbf{Cascaded methods}: OCR+MT+T2I and Qwen3-VL-8B~\cite{bai2025qwen3}+T2I, which decompose the task into sequential text extraction, machine translation, and text rendering stages. 
(2) \textbf{Open-source image editing methods}: FireRed-Image-Edit~\cite{zhou2025fireedit}, LongCat-Image-Edit~\cite{team2025longcat}, and Qwen-Image-Edit~\cite{wu2025qwenimage}. 
(3) \textbf{Closed-source image editing methods}: Seedream~5.0~\cite{seedream2025seedream}, Nano Banana~2~\cite{gemini2026nano}, and GPT Image~2~\cite{openai2026gptimage2}. 
(4) \textbf{Code-driven methods}: approaches that remove text via inpainting, generate structured HTML with VLMs, and render the final image, including Qwen3.5-27B~\cite{team2026qwen35}, GPT5.5~\cite{openai2026gpt55}, and Gemini-3.1 Pro~\cite{gemini2026pro3}.

\subsection{Experimental Results}

We evaluate all methods on TransAnyBench spanning 10 languages, with aggregated results reported in Table~\ref{tab:multilingual}. TransAnyText achieves the best translation quality among open-source methods and outperforms closed-source systems in most language directions. Its explicit modeling of font, color, and positional attributes brings clear advantages in visual fidelity over open-source pixel-based methods, while remaining competitive with closed-source models. For image realism, TransAnyText achieves comparable performance to closed-source systems. As shown in Figure~\ref{fig:teaser} and Figure~\ref{fig:visual_compare}, TransAnyText consistently preserves visual identity and translation accuracy across diverse language pairs. Open-source image editing models often suffer from text rendering errors, while closed-source systems achieve stronger visual quality but may still exhibit style drift and limited editability. By decoupling semantic translation from pixel rendering, TransAnyText achieves superior visual consistency and translation performance.

\noindent\textbf{Analysis of cascaded methods.} Benefiting from dedicated translation modules, cascaded pipelines achieve moderate COMET scores, reflecting reasonable level of cross-lingual performance. However, the erase-and-paste strategy often introduces artifacts on complex backgrounds, leading to significant degradation in visual fidelity and overall realism. In addition, text-to-image components often struggle with multi-region rewriting, which becomes a key bottleneck for further improvement. These observations highlight an inherent limitation of the cascaded paradigm, where optimizing individual components cannot fundamentally prevent error accumulation and propagation across stages.

\noindent\textbf{Analysis of image editing models.} FireRed-Image-Edit, LongCat-Image-Edit, and Qwen-Image-Edit achieve T.Q. scores below 2.9 across translation directions, indicating limited cross-lingual capability. Designed for general-purpose image manipulation, these models struggle to jointly handle visual understanding, translation, and text rendering, resulting in poor character-level accuracy. Despite this, LongCat-Image-Edit and Qwen-Image-Edit maintain moderate visual fidelity (V.F. $>$ 6.0), suggesting that their strengths lie in background synthesis rather than precise text generation. More broadly, the pixel-level end-to-end paradigm entangles semantic translation and visual rendering within a single network, causing objective interference. Closed-source models demonstrate strong performance in translation quality and realism; however, without explicit disentanglement of structure and appearance, their preservation of visual identity remains inconsistent. In addition, their rasterized outputs lack editability, limiting downstream operations such as font substitution or layout adjustment.

\noindent\textbf{Analysis of code-driven methods.} Code-driven methods adopt a paradigm similar to ours, where a VLM generates structured representations followed by background inpainting and deterministic rendering. With Gemini-3.1 Pro as the VLM backbone, this approach achieves competitive visual fidelity comparable to GPT Image~2 while maintaining strong translation quality. These results further validate the effectiveness of decoupling semantic reasoning from pixel synthesis: the VLM focuses on cross-lingual understanding and layout structuring, while text accuracy is ensured by deterministic rendering.

\subsection{Ablation Study}

To assess the contribution of each training stage, we perform a systematic ablation on the mean performance across all translation directions in TransAnyBench, with results reported in Table~\ref{tab:ablation}. The structured HTML representation enables direct computation of fine-grained metrics, allowing performance gains to be explicitly attributed to individual stages. We consider five evaluation dimensions: COMET for translation quality, and T.E. (Text Extraction), T.Q. (Translation Quality), P.A. (Position Accuracy, measured by IoU), and S.S. (Style Similarity) to capture complementary aspects of structured generation.

\noindent\textbf{Analysis of the SFT Stage.} Without task-specific supervision, the base model attains only 0.482 COMET and 0.218 P.A., reflecting a limited capability for image-to-HTML mapping. SFT yields the largest single-stage gain, improving COMET by +0.182 and P.A. by +0.385, thereby establishing the fundamental competence for structured generation.

\noindent\textbf{Analysis of the PWSD Stage.} Compared with vanilla on-policy self-distillation (OPSD)~\cite{zhao2026self}, PWSD achieves more consistent improvements across all metrics. The gains are particularly pronounced in position and style, suggesting that privilege-gap weighting provides more effective supervision for tokens associated with visual attributes. These tokens exhibit weaker structural dependencies in autoregressive generation and therefore benefit more from adaptive reweighting.

\noindent\textbf{Analysis of the GRPO Stage.} Applying GRPO directly after SFT (+SFT+GRPO-only) achieves a COMET score of 0.691, while the full pipeline (+SFT+PWSD+GRPO) further improves performance to 0.704. The gains are particularly evident in P.A. and S.S., indicating that PWSD provides a stronger initialization for subsequent reward optimization. Although GRPO can improve translation and text extraction through task-level reward signals, the lack of dense token-level supervision limits its ability to optimize fine-grained spatial and stylistic attributes. By providing targeted supervision for these under-optimized tokens, PWSD complements GRPO and leads to the best overall performance.

\section{Conclusion}

\begin{table}[t]
\vspace{-2mm}

\centering
\resizebox{\columnwidth}{!}{%
\begin{tabular}{l ccccc}
\toprule
\textbf{Model} & COMET & T.E. & T.Q. & P.A. & S.S. \\
\midrule
Base Model & 0.482 & 0.445 & 5.08 & 0.218 & 0.255 \\
+SFT & 0.664 & 0.877 & 8.59 & 0.603 & 0.554 \\
\midrule
+SFT+OPSD & 0.672 & 0.881 & 8.62 & 0.625 & 0.576 \\
+SFT+PWSD & 0.686 & 0.914 & 8.74 & 0.646 & 0.584 \\
\midrule
+SFT+GRPO-only & 0.691 & 0.928 & 8.94 & 0.667 & 0.601 \\
+SFT+PWSD+GRPO & \textbf{0.704} & \textbf{0.942} & \textbf{8.98} & \textbf{0.697} & \textbf{0.633} \\
\bottomrule
\end{tabular}%
}
\vspace{-1mm}
\caption{Ablation study on each training stages.}
\vspace{-3mm}
\label{tab:ablation}
\end{table}

We formulate e-commerce image text translation as structured visual code generation, where the model outputs a renderable HTML patch instead of pixels. This design separates semantic reasoning from visual rendering: the VLM handles understanding, translation, and structure; the diffusion model supports background inpainting; and a deterministic renderer ensures text accuracy. Based on this formulation, TransAnyText introduces a three-stage post-training pipeline. SFT establishes structured generation, PWSD improves supervision on visually grounded tokens through adaptive weighting, and GRPO refines task performance via multi-dimensional rewards. Experiments on TransAnyBench demonstrate consistent gains over cascaded and end-to-end methods, while maintaining strong editability and controllability.

\noindent\textbf{Limitation.}
While effective for regular layouts, code-based rendering via HTML/CSS has limited expressiveness for complex patterns such as curved or highly stylized text. An optional diffusion refinement module enhances these details, partially alleviating this limitation in practice.

\bibliography{aaai2027}


\end{document}